\documentclass[10pt, a4paper]{article}

\usepackage{booktabs}
\usepackage{latexsym}
\usepackage{xcolor}
\usepackage{comment}
\usepackage[hyphens]{url}
\usepackage{xurl}
\usepackage[final]{lrec2026} 
\usepackage{multirow}
\usepackage{float} 
\usepackage{tcolorbox}
\usepackage{hyperref}
\usepackage{cleveref}

\title{MultiGraSCCo: A Multilingual Anonymization Benchmark with Annotations of Personal Identifiers}

\name{Ibrahim Baroud$^{1,3}$, Christoph Otto$^{1,5}$, Vera Czehmann$^{1,3}$, Christine Hovhannisyan$^{4}$, \\ \large\bf{Lisa Raithel$^{1,2,3,6}$, Sebastian Möller$^{1,3}$, Roland Roller$^{1}$}}

\address{$^{1}$German Research Center for Articial Intelligence (DFKI),\\$^{2}$Charité -- Universitätsmedizin Berlin, Institute for Artificial Intelligence in Medicine, \\$^{3}$Technische Universität Berlin, $^{4}$Humboldt University Berlin, $^{5}$University of Potsdam, $^{6}$BIFOLD \\
         Berlin, Germany \\
         \{ibrahim.baroud, raithel, sebastian.moeller\}@tu-berlin.de\\ 
         \{christoph.otto, vera.czehmann, roland.roller\}@dfki.de\\ 
         christine.hovhannisyan@student.hu-berlin.de\\}

\abstract{
Accessing sensitive patient data for machine learning is challenging due to privacy concerns. 
Datasets with annotations of personally identifiable information are crucial for developing and testing anonymization systems to enable safe data sharing that complies with privacy regulations. 
Since accessing real patient data is a bottleneck, synthetic data offers an efficient solution for data scarcity, bypassing privacy regulations that apply to real data.
Moreover, neural machine translation can help to create high-quality data for low-resource languages by translating validated real or synthetic data from a high-resource language.
In this work, we create a multilingual anonymization benchmark in ten languages, using a machine translation methodology that preserves the original annotations and renders names of cities and people in a culturally and contextually appropriate form in each target language.
Our evaluation study with medical professionals confirms the quality of the translations, both in general and with respect to the translation and adaptation of personal information.
Our benchmark with over 2,500 annotations of personal information can be used in many applications, including training annotators, validating annotations across institutions without legal complications, and helping improve the performance of automatic personal information detection. 
We make our benchmark and annotation guidelines available for further research. 
 \\ \newline  }

\begin{document}

\maketitleabstract

\section{Introduction}
\begin{figure*}[t!]
\centering
\includegraphics[width=\textwidth]{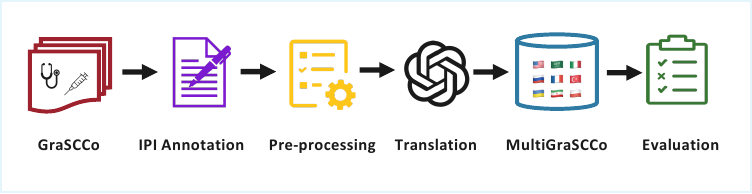}
\vspace{-30pt}
\caption{Our methodology for creating MultiGraSCCo.
We begin by annotating IPIs in the dataset to support researchers in building anonymization systems which can be compliant with European privacy regulations. 
After preprocessing, we use Machine Translation (MT) to translate the texts, together with annotations of direct and indirect identifiers, into 9 languages.
Next, we conduct an evaluation study to assess the quality of the translations. 
Finally, to assess how well privacy-preserving models generalize across languages, we conduct i) monolingual experiments for language-specific performance, ii) cross-lingual experiments to test transfer learning, and iii) multilingual experiments to investigate the benefits of training on the original data in German as well as the translations in one of the target languages.}
\label{fig:methodolgy}
\end{figure*}
Hospitals and medical centers continuously produce huge amounts of data. 
However, data privacy regulations can be a challenge for medical research, as data sources prefer to err on the side of caution to protect patient privacy.
As a result, accessible clinical data is scarce. Even more rare is data in which all personal information is annotated, a necessary step for developing privacy-enhancing technologies.
This has led to a dearth of privacy research and comparatively little development of anonymization and de-identification systems in languages other than English.

Data privacy regulations vary by jurisdiction, making it challenging to create multilingual, multicultural datasets that are privacy-compliant globally. 
For example, in the United States, the Health Insurance Portability and Accountability Act (HIPAA)\footnote{\url{https://www.hhs.gov/hipaa/} (accessed: October 5, 2025)} requires the removal of 18 Protected Health Information (PHI) categories for de-identification, including information such as names, addresses, identifying numbers, and dates. 
On the other hand, the European General Data Protection Regulation (GDPR)\footnote{\url{https://eur-lex.europa.eu/legal-content/en/TXT/?uri=celex\%3A32016R0679} (accessed: October 5, 2025)} does not specify categories for de-identification, but rather demands proof that re-identification is not reasonably possible considering available tools, costs, and time required to re-identify an individual.
This lack of clarity about what constitutes sufficient anonymization makes it challenging to determine whether anonymity has been achieved. 
Additionally, unclear regulations may make data providers hesitant to share data due to concerns about breaking the law.
Nevertheless, given that patient privacy is an ethical and legal priority, a robust anonymization should consider information beyond the scope of HIPAA. 
Personal information ranging from demographics to lifestyle is often left intact after de-identification, leaving the door open for re-identification, especially when multiple attributes are combined \cite{sweeney2002k}. 
Termed quasi-identifiers or indirect personal identifiers (IPIs), these attributes in the context of clinical data are discussed in several works, covering information about family, time of treatment, socioeconomic or criminal history of a patient, and beyond \cite{kolditz2019annotating, feder2020active,baroud-etal-2025-beyond}.
Motivated by the scarcity of datasets containing annotations of personal identifiers, we create a publicly available, multilingual anonymization benchmark built upon the German-language dataset Graz Synthetic Clinical text Corpus (GraSCCo) with PHI annotations \cite{modersohn2022grascco, lohr2024identifying}.
We expand the PHI annotations in GraSCCo with annotations of IPIs. 
\Cref{fig:methodolgy} presents the methodology we used for creating the multilingual anonymization benchmark. 
MultiGraSCCo includes the following language families/languages: German, English (Germanic),
Italian, French (Romance), Arabic (Semitic), Polish, Russian, Ukrainian (Slavic), Turkish
(Turkic), and Persian (Indo-Iranian). 
The writing systems covered are Arabic, Latin, and Cyrillic.
We select a set of languages featuring a diverse range of linguistic phenomena such as inflection, compounding, and subject-object-verb order and for which we were able to find medical professionals who were fluent in both German and one of the target languages.
Our work contributes: 
\begin{itemize}
\item A multilingual anonymization benchmark using machine translation in 10 languages from 6 language families and 3 writing systems, with annotations of direct and indirect personal information;
\item An assessment of the quality of the translations as well as the adaptation of names to the cultural context of the target languages and countries;
\item Monolingual, cross-lingual, and multilingual experiments on the multilingual benchmark to assess usability for model training.
\end{itemize}
Our benchmark can be used as a resource for improving the detection of personal information and training annotators across institutions without legal obstacles, as it contains no real personal information. 
The benchmark and the annotation guidelines can be found on Zenodo.\footnote{\url{https://zenodo.org/records/18847836}}
The best-performing models can be downloaded from Hugging Face.\footnote{\url{https://huggingface.co/collections/Ibrahimbaroud/multilingual-personal-information-detection}}

\section{Related Work}

\paragraph{De-identification} In the context of de-identification, multiple datasets exist. 
One of the most prominent is the 2014 English-language Informatics for Integrating Biology \& the Bedside $($i2b2$)$ data \cite{stubbs2015annotating}, which builds on MIMIC III \cite{johnson2016mimic}. 
A handful of studies also provide resources and systems for the automatic de-identification of textual data in languages other than English.
\citet{catelli2020crosslingual} developed the Italian Society of Radiology (SIRM) COVID-19 de-identification corpus in the Italian language and tested the performance of different techniques in de-identifying medical records in Italian.
For Spanish, \citet{marimon2019automatic} created the MEDDOCAN benchmark for clinical text de-identification.
Moreover, \citet{dalianis2010identifying} created manually annotated, gold-standard benchmarks for the de-identification of Swedish clinical texts. Their corpus was based on a subset of the Stockholm Electronic Patient Records (EPR) Corpus \cite{eprcorpus}.
For German, \citet{modersohn2022grascco} manually developed a small synthetic corpus which was annotated later for Personally Identifiable Information (PII) by \citet{lohr2024identifying}.

\paragraph{Indirect Personal Identifiers (IPIs)} 
Several studies highlight that seemingly non-revealing information, such as ZIP code and gender, can be used in combination to re-identify individuals. Therefore, such information must be considered during anonymization to protect privacy. 
\citet{sweeney2002k} was able to identify Massachusetts governor William Weld's hospital visits and medical information by purchasing the voter registration list for Cambridge, Massachusetts and joining it with a supposedly ``de-identified'' health dataset based on birth date, ZIP code, and gender.
\citet{kolditz2019annotating} and \citet{feder2020active} have suggested adding indirect personal information to the categories provided by HIPAA for robust anonymization. 
Their focus is information such as medical facilities and demographic traits (e.g. marital status, occupation, ethnicity). 
Moreover, \citet{baroud-etal-2025-beyond} have presented a schema of IPIs in clinical texts.
This schema covers a wide variety of information such as family information and criminal or socioeconomic history of the patient which could indirectly contribute to the reidentification of individuals. 
These categories are: \textsc{Family}, \textsc{Appearance}, \textsc{Circumstances}, \textsc{Socioeconomic and Criminal History} , \textsc{Healthcare Facilities and Personnel}, \textsc{Time}, \textsc{Hobbies and Lifestyle}, \textsc{Details about a Direct Identifier} and \textsc{Other}. 

\paragraph{Machine Translation (MT)} There is a significant number of works that focus on translating text resources into other languages. 
\citet{neves-etal-2023-findings} showed the effectiveness of Machine Translation (MT) systems in translating biomedical texts.
Multiple works illustrate that translation combined with large language models can effectively expand resources for various NLP tasks; see \citet{li-callison-burch-2023-paxqa}, \citet{liu-etal-2019-xqa}, and \citet{lewis-etal-2020-mlqa} for Question Answering (QA), \citet{nuutinen-etal-2025-finnish} and \citet{Tulajiang2025ABL} for Span Annotation and Named Entity Recognition (NER), \citet{daza-frank-2020-x} for Semantic Role Labeling (SRL), and \citet{hennig-etal-2023-multitacred} for Relation Extraction (RE). 

\paragraph{Annotation-Preserving MT} Numerous studies have explored the use of MT to address the scarcity of annotated datasets in various languages. 
For example, \citet{seinen2024annotation} used Google Translate and OpenAI GPT-4.1 to translate three corpora from English to Dutch while preserving a wide variety of clinical annotations. 
Their work showed the effectiveness of MT in generating annotation-preserving translations and demonstrated the comparable usability of the Dutch translations to the original data in English for various clinical concept extraction tasks. 
With rare exceptions, the use of MT to address the scarcity of de-identification and privacy-related datasets remains largely unexplored. 
For example, \citet{kocaman-etal-2023-automated} used MT to translate the i2b2 2014 de-identification dataset from English to Arabic and created a de-identification system using the translated dataset, achieving promising results. Their pipeline (BERT For Token Classification (BFTC) + a contextual parser) achieved micro F1-scores ranging from 0.94 to 0.98 on the translated test dataset. 
In their work, entities such as names or age were replaced with their types (e.g. NAME or AGE) in the original texts and then replaced later on in the translation with native Arabic values.

\paragraph{GraSCCo} Unlike many other published clinical datasets, the Graz Synthetic Clinical text Corpus (GraSCCo)\footnote{\url{https://zenodo.org/records/6539131}} is freely accessible without any requirements or training \cite{modersohn2022grascco}. 
The German‑language clinical text corpus contains 63 clinical texts with more than 43,000 tokens covering a large variety of medical topics, such as neurology, oncology, and intensive care. These clinical texts originate from various sources such as hospitals, open access journals, and other texts published on the web.
Each document in the dataset was manually anonymized by a medical doctor to diminish the possibility of re-identifying any persons mentioned in these texts. 
The thorough manual anonymization included shifting dates into the future and replacing real information like names, gender, and locations with fictional ones. 
Additionally, some paragraphs were paraphrased and exchanged across the document; in addition, factual changes were made to medically relevant information, such as diagnoses and test values. 
This rigorous anonymization enabled these documents to be shared without any legal restrictions. 
Despite substantial changes, the anonymized data was found to reflect common clinical language use in a comparison with other real, non-shareable clinical corpora.

\paragraph{GraSCCo\_PHI} \citet{lohr2024identifying} annotated GraSCCo with the 18 PHI categories from HIPAA with the addition of the category \textsc{Profession}, as they considered it  potentially sensitive information. 
The final dataset has been released as open-source and contains 1,438 PHI annotations, indicating that approximately 3\% of the tokens in the corpus correspond to personal information according to HIPAA.
The authors report an inter-annotator agreement of Krippendorff's $\alpha \approx 0.97$.

\section{Annotating IPIs in GraSCCo} 

Motivated by the work of \citet{kolditz2019annotating}, \citet{feder2020active}, and \citet{baroud-etal-2025-beyond}, we annotate IPIs in GraSCCo to enable privacy models to be trained to detect information beyond classical PHI categories that may result in the re-identification of patients.
Our IPI annotations can help anonymization models to be compliant with European privacy regulations, since mere de-identification as described in HIPAA may not be sufficient.   
The following sections introduce the IPI categories we annotated, our adaptation of the IPI annotation guidelines to this German dataset, and the setup of the annotation task. 
Moreover, we present statistics on the IPI annotations added and inter-annotator agreement. 

\subsection{Indirect Personal Identifiers (IPI) Schema}\label{ipi_schema}

We adapted the guidelines from \citet{baroud-etal-2025-beyond} for the German-language GraSCCo corpus.
Annotator feedback was used in order to enhance the clarity and generalizability of the guidelines. 
The final schema used in this work has 13 categories. 
We introduced the categories: \textsc{Ethnicity}, \textsc{Languages and speech}, and \textsc{Sexual orientation}.
Moreover, the category \textsc{Details about location} was added to narrow the scope of the broader category \textsc{Details about a Direct Identifier} and to improve the automatic detection of location-specific information.
Finally, any other indirect information was annotated under \textsc{Other}. 

\subsection{The IPI Annotation Task}

Annotation was conducted using INCEpTION,\footnote{\url{https://inception-project.github.io/}, version 35.2.} an open-source web-based text annotation platform which supports collaborative annotation \cite{tubiblio106270}.
The annotators labeled all IPIs in the GraSCCo documents as defined in \Cref{ipi_schema}. 
The annotation process was conducted in multiple phases. 
In each phase, 15 documents were annotated by both annotators, then discussed with the first author to answer questions and resolve ambiguities. Finally, both annotators and the first author consolidated both sets of annotations into one final version.

\subsection{IPI Characteristics in GraSCCo}

We calculated the inter-annotator agreement (IAA) using the average pairwise relaxed F1-score as in \citet{hripcsak2005agreement}.
Following this schema, a true positive is achieved when both annotations overlap and have the same label.
The average pairwise relaxed F1-score is more suitable for span annotation than other scores such as the Krippendorff's $\alpha$, since the number of negative examples is very high and the probability of chance agreement on the desired spans is close to zero \cite{hripcsak2005agreement}.
\Cref{table:ipis} summarizes the overall and per-category number of spans annotated using the IPI schema. 
The categories \textsc{Time} (47.89\%) and \textsc{Fclt\_Personnel} (34.21\%) comprise the highest proportion of the annotations.
We record an IAA micro F1-score of 83\% (macro 72\%).\footnote{We do not include the category \textsc{OTHER} in this calculation because it was merely added as a backup category for miscellaneous information and does not represent any specific kind of IPI.}
\Cref{table:ipis} presents the IAA per category and overall. 

\begin{table}
\small
\centering
\begin{tabular}{l|l|l|l}
\hline
\textbf{\begin{tabular}[c]{@{}l@{}}IPI \\ Category\end{tabular}}                                                & \textbf{Count} & \textbf{Percent.} & \textbf{\begin{tabular}[c]{@{}l@{}}Inter-\\ annotator \\ agreement\end{tabular}} \\ \hline
\textsc{Time}             & 546            & 47.89\%             & 0.85                                                                          \\ \hline
\textsc{Fclt\_Personnel}            & 390            & 34.21\%             & 0.86                                                                          \\ \hline
\textsc{Appearance}           & 64             & 5.61\%              & 0.79                                                                          \\ \hline
\textsc{Circumstances}                                                             & 64             & 5.61\%              & 0.61                                                                          \\ \hline
\textsc{Family}                                                               & 37             & 3.25\%              & 0.85                                                                          \\ \hline
\begin{tabular}[c]{@{}l@{}}\textsc{Hobbies and}\\ \textsc{Lifestyle}\end{tabular}       & 14             & 1.23\%              & 0.58                                                                          \\ \hline
\begin{tabular}[c]{@{}l@{}}\textsc{Socio-economic and}\\ \textsc{criminal history}\\ \textsc{(SEC)}\end{tabular}                                                                  & 14             & 1.23\%              & 0.33                                                                          \\ \hline
\begin{tabular}[c]{@{}l@{}}\textsc{Details about a} \\ \textsc{Direct Identifier}\end{tabular} & 0              & 0.0\%              & -                                                                          \\ \hline
\begin{tabular}[c]{@{}l@{}}\textsc{Details about} \\ \textsc{Location}\end{tabular} & 6              & 0.53\%              & 0.57                                                                          \\ \hline
\textsc{Languages}                                                            & 3              & 0.26\%              & 0.80                                                                           \\ \hline
\begin{tabular}[c]{@{}l@{}}\textsc{Sexual}\\ \textsc{Orientation}\end{tabular}         & 1              & 0.09\%              & 1.0                                                                           \\ \hline
\textsc{Ethnicity}                                                                & 0              & 0.0\%              & -                                                                          \\ \hline
\textsc{Other}                                                                & 1              & 0.09\%              & 0.17                                                                          \\ \hline
All                                                                  & 1140           &                     & \shortstack{micro: 0.83 \\ macro: 0.73}                                                                          \\ \hline
\end{tabular}
\caption{IPI annotation counts, percentages from overall IPI annotations, and inter-annotator agreement (IAA).}
\label{table:ipis}
\end{table}

\section{Translating GraSCCo}
Following the annotation process, we translated the German dataset into English, French, Arabic, Persian, Italian, Polish, Russian, Ukrainian, and Turkish.
This section introduces our methodology for data preprocessing, translation, and quality analysis. 

\subsection{Preprocessing}

\begin{figure*}[ht]
\centering
\includegraphics[width=\textwidth]{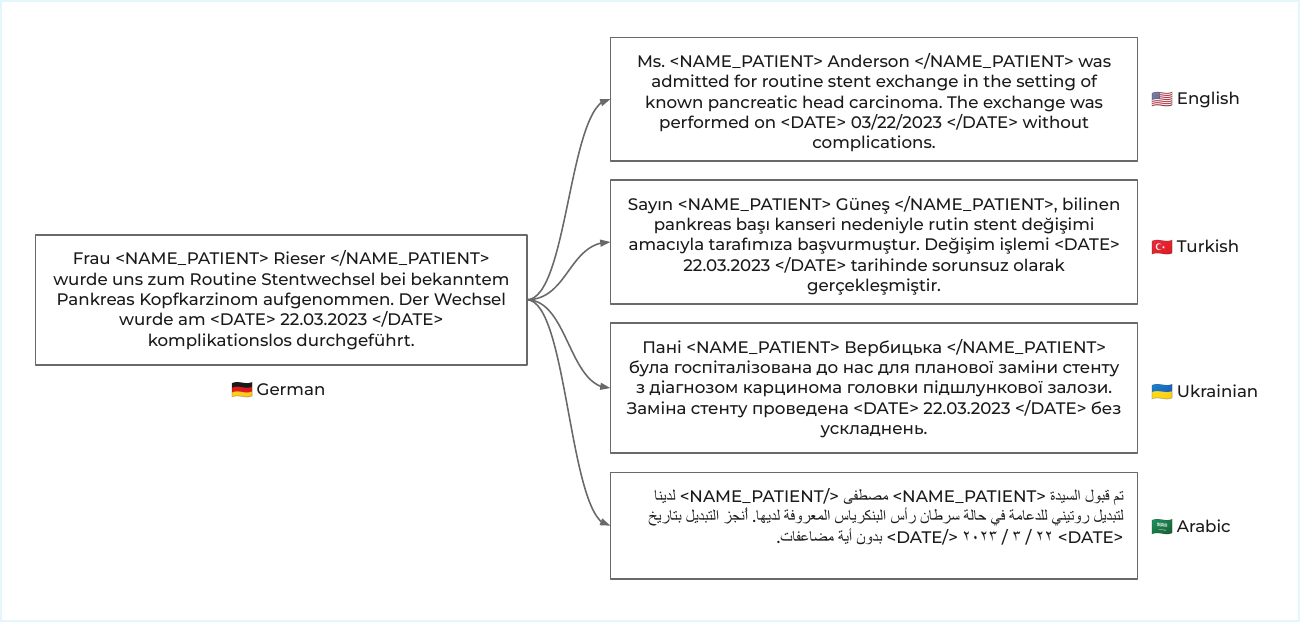}
\vspace{-15pt}
\caption{Sample translations from German to English, Turkish, Ukrainian, and Arabic with XML tags around the start and end of annotated spans.}
\label{fig:translation_examples}
\end{figure*}

\paragraph{Tagging PHI and IPI Information} In this preprocessing step, start and end XML tags were added around all PHI and IPI annotations to ensure valid annotation mapping in the target languages after performing the automatic translation.
For example, as shown in \Cref{fig:translation_examples}, the text span, ``The patient's name is John Smith'' becomes, ``The patient's name is <NAME\_PATIENT>John Smith</NAME\_PATIENT>.'' 

\paragraph{Typo Correction and Abbreviation Expansion} In GraSCCo, typos were intentionally introduced in the documents to mimic real clinical data and provide a more realistic challenge for training NLP models. 
Moreover, GraSCCo, like most clinical data, is full of medical abbreviations, which can introduce ambiguity and influence the quality of the translations. 
For the purposes of our translation study, we therefore asked our annotators to label all typos and abbreviations in the documents during the annotation process. 
Next, we submitted these annotations together with the original documents to GPT-4.1 to correct the typos and expand the abbreviations.
The preprocessing prompt can be found in \Cref{fig:preprocess_prompt} in \Cref{sec:preprocess_eval_sec}.
Finally, to ensure the quality of preprocessing, both annotators evaluated a sample of 15 documents (24\% of GraSCCo) for the model's ability to correct the typos, expand the abbreviations correctly, and keep the semantic and syntactic features of the original documents without introducing hallucinations. 
The manual evaluation showed that the model was able to correct all typos and preserve the XML tag structure as introduced in the previous paragraph.
The model was also able to expand over 96\% of the annotated abbreviations correctly. 
Incorrect expansions were corrected by the annotators and included in the final dataset. More details about the manual evaluation of the preprocessing can be found in \Cref{sec:preprocess_eval_sec}.

\subsection{Translation using GPT-4.1} The OpenAI API\footnote{\url{https ://openai.com/api/}} was used to translate the original clinical documents into the various languages. 
The translation prompt can be found in \Cref{sec:aut_trans}. 
We prompted the model not only to translate the documents literally, but also to adapt names (of people, streets, cities, or institutions) into the target language to ensure they are culturally and contextually appropriate in the target country, as shown in \Cref{fig:translation_examples}. 
This method renders a more natural and useful text than a literal translation,  especially given the findings of \citet{kocaman-etal-2023-automated} that anonymization models trained on contextually correct translations generalize better when tested on real data.
We set a target country for each target language to avoid inconsistencies such as variable date formatting.

\subsection{Translation Quality Analysis}

To evaluate the quality of the machine translation from the original German documents into the target languages, we conducted a human evaluation study. 
10 sample documents per language were assessed. 
Each document was evaluated by at least two native speakers of the target language who were either physicians or medical students and could also speak German fluently. 
Participants were paid €10 per evaluated document. 
To ensure more fine-grained and higher-quality evaluations, we split long documents into chunks of 150-200 words and later averaged the ratings of each chunk to generate a score for the whole document. 
Entities such as names and addresses which needed to be culturally adapted in the translation were marked using XML tags in the original text and the translations, as shown in \Cref{fig:translation_examples}.

The translations were rated along four dimensions on a Likert scale from 1 to 7, similarly to the methodology outlined in \citet{Kocmi2022FindingsOT}: \textit{General Quality}, \textit{Grammar}, \textit{Tag Structure}, and \textit{Entity Translation Quality}. 
To make the rating scale clear to the participants, we added the following labels to the possible responses, as proposed by \citet{xu2024contrastive}: \textit{1 (nonsensical)}, \textit{3 (partially correct)}, \textit{5 (largely correct)}, \textit{7 (perfect)}. 
The following four prompts were used in the study: 
\begin{enumerate}
        \item \textit{General Quality:} How well does the translated text preserve the semantic meaning of the original, despite minor translation errors?
        \item \textit{Grammar:} How would you rate the overall grammatical correctness of the translation?
        \item \textit{Tag Structure:} In the translated text, the tags were kept intact without interference, regardless of the text between the start and end tags.\footnote{For clarity, the responses for this prompt were labeled \textit{1 (totally disagree)} to \textit{7 (totally agree)}.}
        \item \textit{Entity Translation Quality:} How would you rate the translation of the text between tags (e.g., names, locations, cities) in terms of being adapted culturally, contextually, and linguistically to the target language, rather than directly translated?
\end{enumerate}
After rating the four dimensions, evaluators were given the opportunity to add comments and make corrections to the translation.

\section{Experiments}

We explored the task of multilingual de-identification through three baseline experiments  that simulate different levels of resource availability: i) monolingual, ii) cross-lingual, and ii) multilingual settings. As the German GraSSCo corpus is the starting point for our multilingual dataset, we refer to German as the source language for all experiments. In the monolingual setup, a separate model was trained and evaluated on each language using its own data, representing the baseline high-resource scenario. The cross-lingual setup modeled zero-shot performance by training a model only on the German source data before evaluating it on the test sets of all other languages. This replicates a scenario in which no labeled data exists in the target language. Finally, the multilingual setup combined German training data with a limited amount of annotated data from the target language. For each language, we experimented with different proportions of in-language training data (25\%, 50\%, 75\% and 100\%) to observe how performance scaled with additional supervision. Together, these setups allowed us to assess language-specific baselines, cross-lingual generalization, and the benefit of additional in-language data.

\begin{table*}
\small
\centering
\begin{tabular}{l|cccccc}
\hline
\textbf{Language} & \textbf{All Micro F1} & \textbf{All Macro F1} & \textbf{PHI Micro F1} & \textbf{PHI Macro F1} & \textbf{IPI Micro F1} & \textbf{IPI Macro F1} \\ \hline
English   & 0.870 \textcolor{gray}{\scriptsize (0.850)} & 0.759 \textcolor{gray}{\scriptsize (0.710)} & 0.933 \textcolor{gray}{\scriptsize (0.923)} & 0.858 \textcolor{gray}{\scriptsize (0.831)} & 0.799 \textcolor{gray}{\scriptsize (0.770)} & 0.611 \textcolor{gray}{\scriptsize (0.518)} \\
French    & 0.833 \textcolor{gray}{\scriptsize (0.837)} & 0.685 \textcolor{gray}{\scriptsize (0.768)} & 0.927 \textcolor{gray}{\scriptsize (0.928)} & 0.832 \textcolor{gray}{\scriptsize (0.888)} & 0.733 \textcolor{gray}{\scriptsize (0.740)} & 0.465 \textcolor{gray}{\scriptsize (0.588)} \\
German    & 0.874 \textcolor{gray}{\scriptsize (0.844)} & 0.804 \textcolor{gray}{\scriptsize (0.732)} & 0.955 \textcolor{gray}{\scriptsize (0.941)} & 0.888 \textcolor{gray}{\scriptsize (0.847)} & 0.785 \textcolor{gray}{\scriptsize (0.736)} & 0.676 \textcolor{gray}{\scriptsize (0.559)} \\
Arabic    & 0.873 \textcolor{gray}{\scriptsize (0.845)} & 0.794 \textcolor{gray}{\scriptsize (0.726)} & 0.951 \textcolor{gray}{\scriptsize (0.932)} & 0.906 \textcolor{gray}{\scriptsize (0.843)} & 0.787 \textcolor{gray}{\scriptsize (0.751)} & 0.625 \textcolor{gray}{\scriptsize (0.550)} \\
Persian   & 0.864 \textcolor{gray}{\scriptsize (0.844)} & 0.772 \textcolor{gray}{\scriptsize (0.745)} & 0.944 \textcolor{gray}{\scriptsize (0.935)} & 0.888 \textcolor{gray}{\scriptsize (0.862)} & 0.778 \textcolor{gray}{\scriptsize (0.752)} & 0.598 \textcolor{gray}{\scriptsize (0.557)} \\
Italian   & 0.861 \textcolor{gray}{\scriptsize (0.850)} & 0.781 \textcolor{gray}{\scriptsize (0.758)} & 0.938 \textcolor{gray}{\scriptsize (0.930)} & 0.901 \textcolor{gray}{\scriptsize (0.882)} & 0.776 \textcolor{gray}{\scriptsize (0.761)} & 0.601 \textcolor{gray}{\scriptsize (0.597)} \\
Polish    & 0.866 \textcolor{gray}{\scriptsize (0.839)} & 0.760 \textcolor{gray}{\scriptsize (0.724)} & 0.931 \textcolor{gray}{\scriptsize (0.918)} & 0.855 \textcolor{gray}{\scriptsize (0.823)} & 0.803 \textcolor{gray}{\scriptsize (0.770)} & 0.634 \textcolor{gray}{\scriptsize (0.573)} \\
Russian   & 0.781 \textcolor{gray}{\scriptsize (0.850)} & 0.661 \textcolor{gray}{\scriptsize (0.754)} & 0.865 \textcolor{gray}{\scriptsize (0.927)} & 0.772 \textcolor{gray}{\scriptsize (0.849)} & 0.696 \textcolor{gray}{\scriptsize (0.774)} & 0.494 \textcolor{gray}{\scriptsize (0.617)} \\
Ukrainian & 0.809 \textcolor{gray}{\scriptsize (0.849)} & 0.700 \textcolor{gray}{\scriptsize (0.714)} & 0.900 \textcolor{gray}{\scriptsize (0.921)} & 0.810 \textcolor{gray}{\scriptsize (0.839)} & 0.720 \textcolor{gray}{\scriptsize (0.773)} & 0.534 \textcolor{gray}{\scriptsize (0.527)} \\
Turkish   & 0.728 \textcolor{gray}{\scriptsize (0.848)} & 0.596 \textcolor{gray}{\scriptsize (0.775)} & 0.763 \textcolor{gray}{\scriptsize (0.927)} & 0.642 \textcolor{gray}{\scriptsize (0.870)} & 0.690 \textcolor{gray}{\scriptsize (0.769)} & 0.528 \textcolor{gray}{\scriptsize (0.634)} \\ \hline
\end{tabular}
\caption{Monolingual experiment results across 5 folds (\small\textcolor{gray}{mmBERT results reference}).}
\label{table:monolingual_f1_comparison}
\end{table*}

\begin{table*}
\small
\centering
\begin{tabular}{l|cccccc}
\hline
\textbf{Language} & \textbf{All Micro F1} & \textbf{All Macro F1} & \textbf{PHI Micro F1} & \textbf{PHI Macro F1} & \textbf{IPI Micro F1} & \textbf{IPI Macro F1} \\ \hline
English   & 0.798 \textcolor{gray}{\scriptsize ± 0.012} & 0.703 \textcolor{gray}{\scriptsize ± 0.031} & 0.880 \textcolor{gray}{\scriptsize ± 0.011} & 0.806 \textcolor{gray}{\scriptsize ± 0.025} & 0.706 \textcolor{gray}{\scriptsize ± 0.031} & 0.548 \textcolor{gray}{\scriptsize ± 0.044} \\
French    & 0.768 \textcolor{gray}{\scriptsize ± 0.019} & 0.680 \textcolor{gray}{\scriptsize ± 0.028} & 0.873 \textcolor{gray}{\scriptsize ± 0.021} & 0.804 \textcolor{gray}{\scriptsize ± 0.040} & 0.662 \textcolor{gray}{\scriptsize ± 0.023} & 0.493 \textcolor{gray}{\scriptsize ± 0.015} \\
Arabic    & 0.740 \textcolor{gray}{\scriptsize ± 0.047} & 0.634 \textcolor{gray}{\scriptsize ± 0.040} & 0.860 \textcolor{gray}{\scriptsize ± 0.029} & 0.773 \textcolor{gray}{\scriptsize ± 0.049} & 0.618 \textcolor{gray}{\scriptsize ± 0.060} & 0.424 \textcolor{gray}{\scriptsize ± 0.078} \\
Persian     & 0.720 \textcolor{gray}{\scriptsize ± 0.031} & 0.568 \textcolor{gray}{\scriptsize ± 0.033} & 0.816 \textcolor{gray}{\scriptsize ± 0.050} & 0.635 \textcolor{gray}{\scriptsize ± 0.031} & 0.631 \textcolor{gray}{\scriptsize ± 0.034} & 0.468 \textcolor{gray}{\scriptsize ± 0.042} \\
Italian   & 0.790 \textcolor{gray}{\scriptsize ± 0.017} & 0.686 \textcolor{gray}{\scriptsize ± 0.047} & 0.861 \textcolor{gray}{\scriptsize ± 0.010} & 0.774 \textcolor{gray}{\scriptsize ± 0.028} & 0.709 \textcolor{gray}{\scriptsize ± 0.027} & 0.555 \textcolor{gray}{\scriptsize ± 0.085} \\
Polish    & 0.788 \textcolor{gray}{\scriptsize ± 0.025} & 0.606 \textcolor{gray}{\scriptsize ± 0.048} & 0.839 \textcolor{gray}{\scriptsize ± 0.017} & 0.616 \textcolor{gray}{\scriptsize ± 0.054} & 0.739 \textcolor{gray}{\scriptsize ± 0.035} & 0.591 \textcolor{gray}{\scriptsize ± 0.118} \\
Russian   & 0.742 \textcolor{gray}{\scriptsize ± 0.020} & 0.604 \textcolor{gray}{\scriptsize ± 0.029} & 0.797 \textcolor{gray}{\scriptsize ± 0.026} & 0.672 \textcolor{gray}{\scriptsize ± 0.048} & 0.690 \textcolor{gray}{\scriptsize ± 0.021} & 0.503 \textcolor{gray}{\scriptsize ± 0.041} \\
Ukrainian & 0.724 \textcolor{gray}{\scriptsize ± 0.024} & 0.641 \textcolor{gray}{\scriptsize ± 0.036} & 0.785 \textcolor{gray}{\scriptsize ± 0.039} & 0.720 \textcolor{gray}{\scriptsize ± 0.048} & 0.661 \textcolor{gray}{\scriptsize ± 0.031} & 0.535 \textcolor{gray}{\scriptsize ± 0.036} \\
Turkish   & 0.718 \textcolor{gray}{\scriptsize ± 0.041} & 0.621 \textcolor{gray}{\scriptsize ± 0.053} & 0.780 \textcolor{gray}{\scriptsize ± 0.058} & 0.699 \textcolor{gray}{\scriptsize ± 0.077} & 0.653 \textcolor{gray}{\scriptsize ± 0.048} & 0.504 \textcolor{gray}{\scriptsize ± 0.092} \\ 
\hline
\end{tabular}
\caption{Crosslingual experiment results across 5 folds (\textcolor{gray}{± SD}).}
\label{table:crosslingual_f1}
\end{table*}

\begin{table*}
\small  
\centering
\begin{tabular}{ll|ccccc}
\hline
\textbf{Metric} & \textbf{Language} & \textbf{0\%} & \textbf{25\%} & \textbf{50\%} & \textbf{75\%} & \textbf{100\%} \\ \hline
 & English   & 0.798 \textcolor{gray}{\scriptsize ± 0.012} & 0.843 \textcolor{gray}{\scriptsize ± 0.009} & 0.851 \textcolor{gray}{\scriptsize ± 0.015} & 0.862 \textcolor{gray}{\scriptsize ± 0.009} & 0.867 \textcolor{gray}{\scriptsize ± 0.011} \\
 & French    & 0.768 \textcolor{gray}{\scriptsize ± 0.019} & 0.833 \textcolor{gray}{\scriptsize ± 0.014} & 0.848 \textcolor{gray}{\scriptsize ± 0.012} & 0.853 \textcolor{gray}{\scriptsize ± 0.013} & 0.863 \textcolor{gray}{\scriptsize ± 0.007} \\
 & Arabic    & 0.740 \textcolor{gray}{\scriptsize ± 0.047} & 0.843 \textcolor{gray}{\scriptsize ± 0.016} & 0.851 \textcolor{gray}{\scriptsize ± 0.008} & 0.857 \textcolor{gray}{\scriptsize ± 0.009} & 0.862 \textcolor{gray}{\scriptsize ± 0.019} \\
 & Persian   & 0.720 \textcolor{gray}{\scriptsize ± 0.031} & 0.827 \textcolor{gray}{\scriptsize ± 0.009} & 0.843 \textcolor{gray}{\scriptsize ± 0.008} & 0.856 \textcolor{gray}{\scriptsize ± 0.015} & 0.865 \textcolor{gray}{\scriptsize ± 0.007} \\
Micro F1 & Italian   & 0.790 \textcolor{gray}{\scriptsize ± 0.017} & 0.855 \textcolor{gray}{\scriptsize ± 0.014} & 0.862 \textcolor{gray}{\scriptsize ± 0.012} & 0.869 \textcolor{gray}{\scriptsize ± 0.006} & 0.879 \textcolor{gray}{\scriptsize ± 0.010} \\
 & Polish    & 0.788 \textcolor{gray}{\scriptsize ± 0.025} & 0.841 \textcolor{gray}{\scriptsize ± 0.008} & 0.861 \textcolor{gray}{\scriptsize ± 0.013} & 0.865 \textcolor{gray}{\scriptsize ± 0.006} & 0.874 \textcolor{gray}{\scriptsize ± 0.010} \\
 & Russian   & 0.742 \textcolor{gray}{\scriptsize ± 0.020} & 0.839 \textcolor{gray}{\scriptsize ± 0.009} & 0.845 \textcolor{gray}{\scriptsize ± 0.006} & 0.868 \textcolor{gray}{\scriptsize ± 0.010} & 0.867 \textcolor{gray}{\scriptsize ± 0.004} \\
 & Ukrainian & 0.724 \textcolor{gray}{\scriptsize ± 0.024} & 0.841 \textcolor{gray}{\scriptsize ± 0.009} & 0.856 \textcolor{gray}{\scriptsize ± 0.007} & 0.858 \textcolor{gray}{\scriptsize ± 0.014} & 0.867 \textcolor{gray}{\scriptsize ± 0.013} \\ 
 & Turkish   & 0.718 \textcolor{gray}{\scriptsize ± 0.041} & 0.830 \textcolor{gray}{\scriptsize ± 0.013} & 0.847 \textcolor{gray}{\scriptsize ± 0.014} & 0.856 \textcolor{gray}{\scriptsize ± 0.014} & 0.867 \textcolor{gray}{\scriptsize ± 0.012} \\
\hline
 & English   & 0.703 \textcolor{gray}{\scriptsize ± 0.031} & 0.744 \textcolor{gray}{\scriptsize ± 0.033} & 0.763 \textcolor{gray}{\scriptsize ± 0.017} & 0.788 \textcolor{gray}{\scriptsize ± 0.033} & 0.795 \textcolor{gray}{\scriptsize ± 0.017} \\
 & French    & 0.680 \textcolor{gray}{\scriptsize ± 0.028} & 0.770 \textcolor{gray}{\scriptsize ± 0.030} & 0.790 \textcolor{gray}{\scriptsize ± 0.032} & 0.800 \textcolor{gray}{\scriptsize ± 0.023} & 0.812 \textcolor{gray}{\scriptsize ± 0.013} \\
 & Arabic    & 0.634 \textcolor{gray}{\scriptsize ± 0.040} & 0.776 \textcolor{gray}{\scriptsize ± 0.018} & 0.781 \textcolor{gray}{\scriptsize ± 0.014} & 0.786 \textcolor{gray}{\scriptsize ± 0.027} & 0.763 \textcolor{gray}{\scriptsize ± 0.037} \\
 & Persian   & 0.568 \textcolor{gray}{\scriptsize ± 0.033} & 0.730 \textcolor{gray}{\scriptsize ± 0.029} & 0.746 \textcolor{gray}{\scriptsize ± 0.017} & 0.766 \textcolor{gray}{\scriptsize ± 0.033} & 0.767 \textcolor{gray}{\scriptsize ± 0.019} \\
Macro F1 & Italian   & 0.686 \textcolor{gray}{\scriptsize ± 0.047} & 0.793 \textcolor{gray}{\scriptsize ± 0.023} & 0.796 \textcolor{gray}{\scriptsize ± 0.028} & 0.829 \textcolor{gray}{\scriptsize ± 0.025} & 0.834 \textcolor{gray}{\scriptsize ± 0.012} \\
 & Polish    & 0.606 \textcolor{gray}{\scriptsize ± 0.048} & 0.707 \textcolor{gray}{\scriptsize ± 0.054} & 0.720 \textcolor{gray}{\scriptsize ± 0.033} & 0.782 \textcolor{gray}{\scriptsize ± 0.046} & 0.783 \textcolor{gray}{\scriptsize ± 0.030} \\
 & Russian   & 0.604 \textcolor{gray}{\scriptsize ± 0.029} & 0.739 \textcolor{gray}{\scriptsize ± 0.025} & 0.746 \textcolor{gray}{\scriptsize ± 0.023} & 0.803 \textcolor{gray}{\scriptsize ± 0.027} & 0.806 \textcolor{gray}{\scriptsize ± 0.022} \\
 & Ukrainian & 0.641 \textcolor{gray}{\scriptsize ± 0.036} & 0.745 \textcolor{gray}{\scriptsize ± 0.025} & 0.785 \textcolor{gray}{\scriptsize ± 0.015} & 0.786 \textcolor{gray}{\scriptsize ± 0.018} & 0.799 \textcolor{gray}{\scriptsize ± 0.029} \\
 & Turkish   & 0.621 \textcolor{gray}{\scriptsize ± 0.053} & 0.730 \textcolor{gray}{\scriptsize ± 0.032} & 0.776 \textcolor{gray}{\scriptsize ± 0.047} & 0.782 \textcolor{gray}{\scriptsize ± 0.028} & 0.817 \textcolor{gray}{\scriptsize ± 0.035} \\
\hline
\end{tabular}
\caption{Multilingual experiment results across different training data proportions per language (\small\textcolor{gray}{±SD}).}
\label{table:multilingual_f1_micro_macro_summary}
\end{table*}

\subsection{Experimental Setup} 

All monolingual experiments were conducted using language-specific encoder models based on either RoBERTa-base or BERT-base architecture.\footnote{\footnotesize{de: deepset/gbert-base, en: FacebookAI/roberta-base, ar: aubmindlab/bert-base-arabertv02, fr: almanach/camembert-base, pr: sbunlp/fabert, uk: KoichiYasuoka/roberta-base-ukrainian, ru: blinoff/roberta-base-russian-v0, pl: sdadas/polish-roberta-base-v2, it: osiria/roberta-base-italian, tr: TURKCELL/roberta-base-turkish-uncased}} 
This safeguards a comparable parameter size and performance across different languages. To complement our baseline with a state-of-the-art model, all cross-lingual and multilingual experiments were conducted using multilingual-modernBERT-base (mmBERT) \citep{marone2025mmbertmodernmultilingualencoder}. For reference, we also ran all monolingual experiments with mmBERT. 

Training stability was enhanced by removing labels with 11 or fewer training examples and merging all name label variants (\textsc{Name Patient}, \textsc{Name Doctor}, \textsc{Name Title}, \textsc{Name Relative}, \textsc{Name Username}, \textsc{Name Extern}) into a single class \textsc{Phi-Name}. Documents were segmented at the sentence level and stratified into 65\% training, 15\% validation, and 20\% test splits. Further, we performed a grid search to select the best hyperparameters for each model. For evaluation, we used NERvaluate\footnote{\url{https://github.com/MantisAI/nervaluate}} with relaxed span-level matching mode. This mode accounts for partial span mismatches and reflects the practical requirements of privacy tasks, where high recall is important. All results are reported as the average across five different stratified runs.

\section{Results and Discussion}

In this section, we present insights into the translation quality of MultiGraSCCo and discuss interesting features of the translations in the different languages.
We also present and discuss the performance of the models trained in various setups. 

\subsection{IPI Annotation}

Inter-annotator agreement showed considerable variability across IPI categories (see \Cref{table:ipis}, micro F1=0.83, macro F1=0.73).
Variability is characteristic of span-based annotation, in contrast to simple entity annotation (such as names or numbers), as it requires additional decisions about span length and segmentation. 

High-frequency categories achieved consistently higher agreement.
The two most frequent categories, \textsc{Time} and \textsc{Fclt\_Personnel}, recorded the highest agreement scores of F1=0.85 and F1=0.86 respectively.
These categories were comparatively concrete and lexically anchored, resulting in fewer borderline cases.
By contrast, achieving high agreement for other, more descriptive categories was challenging and required refining the category definitions during the annotation process. 
For instance, \textsc{Appearance} (F1=0.79) was calibrated to cover only characteristics that would be permanently visible to a lay observer, explicitly excluding medical diagnoses even when they could imply physical traits (e.g. obesity).
The lowest agreement was observed for \textsc{Socio-economic and criminal history} (F1=0.33). 
During guideline development, we iteratively refined the definition of this category; one such refinement was to extend coverage to include non-employment roles (e.g., being a student).

In defining the categories, we erred on the side of caution, since a certain phrase or unit of information may be revealing in some contexts and benign in others. 
This sometimes resulted in the over-annotation of non-revealing information; for example, all instances of words such as ``today'' and ``tomorrow'' were removed under the category \textsc{Time}. 
This was done under the assumption that casting a wider net would remove as many potentially sensitive spans as possible without undermining the utility of the data, even if some benign information was also lost in the process.


Finally, only one annotation in the backup category \textsc{Other} was included in the curated dataset: a patient's inability to attend an appointment due to a six-week cruise. 
This suggests that our IPI annotation categories account for almost all types of indirect information that are present in the data.

\subsection{Translation Quality}

\paragraph{Automatic validation of annotations}
\begin{table*}
\small
\centering
\begin{tabular}{@{}lllllllllll@{}}
\toprule
\textbf{Language}       & En     & Fr    & Ar    & Pr    & It    & Pl    & Ru    & Tr    & Uk    & Average \\ \midrule
\textbf{Preserved (\%)} & 100.00 & 99.84 & 99.57 & 99.73 & 99.92 & 99.84 & 99.96 & 98.99 & 99.96 & 99.76   \\ \bottomrule
\end{tabular}
\caption{Percentages of personal information annotations preserved in the translations of each language and overall.}
\label{table:automatic_eval}
\end{table*}

\begin{figure}[!ht]
\begin{center}
\includegraphics[width=\columnwidth]{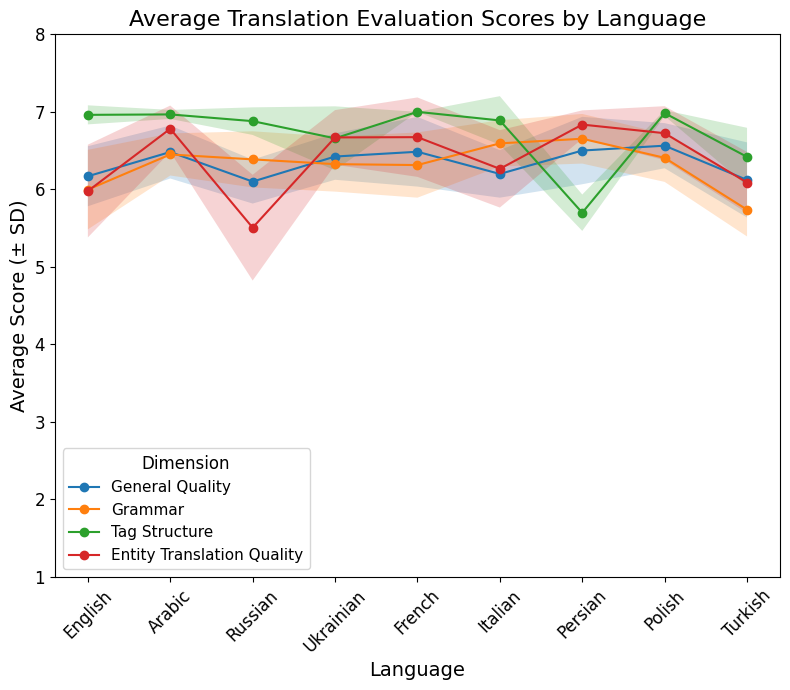}
\vspace{-10pt}
\caption{Results of the qualitative evaluation of the automatic translation into the nine languages. Each dimension was evaluated using a Likert scale between 1 and 7.}
\label{fig:trans_eval_compact}
\end{center}
\end{figure}

Our evaluation study records an average \textit{General Quality} score of 6.34 out of 7 across languages, 6.32 for \textit{Grammar}, 6.72 for \textit{Tag Structure}, and 6.39 for \textit{Entity Translation Quality}. 
\Cref{fig:trans_eval_compact} shows the results of the translation evaluation study.
We calculate Inter-Rater Agreement (IRA) as the Mean Absolute Difference (MAD) \cite{zhan-etal-2023-evaluating} between the ratings. 
We report a MAD of 0.61 across all languages and evaluation dimensions.
This indicates high agreement with only a small difference in rating compared to the rating scale (1-7). 
Moreover, 93.34\% of all differences between rating pairs fall within one point of the rating scale, while 47.22\% show perfect agreement (zero difference). 
More details on the IRA per dimension can be found in \Cref{sec:ira_translations}.
According to evaluators, the overall semantic meaning of the original documents were preserved in the GPT-4.1 translations with a relatively high grammatical quality.
Moreover, the raters found the adaptation of personal information such as names and addresses to the target languages and contexts to be culturally appropriate (average score of 6.39 out of 7 points).
This confirms that state-of-the-art language models can be powerful tools not only for high-quality translations of clinical texts, but also for adapting span annotations to be culturally and contextually relevant rather than being literally translated or replaced using predefined lists.
\paragraph{Translation Error Analysis} Interesting insights about the translations were provided by the raters in the comment section of the evaluation study.
Some documents in English were perceived as less natural due to GPT-4.1's choice of personal and city names in the translations. 
For example, the model sometimes provided two locations with an unlikely distance between them compared to the original text. 
Since English was the first language we evaluated, we improved our prompt further to avoid such errors in other languages and retranslated the incorrect English translations using the final prompt shown in \Cref{sec:aut_trans}.\\ 

Translations into several languages including Arabic, Turkish, and Persian contained errors regarding medical terminology. 
Additionally, in Arabic, the model used Latin terminology for medical conditions or body parts, which, despite being intelligible to physicians, have Arabic equivalents used more commonly by native speakers. 
For example, ``\textit{mitral} valve'' was rendered as ``al-sharayan \textit{al-mitrali}'' instead of the more commonplace ``al-sharayan \textit{at-taji}.''
Moreover, translations into Russian, Turkish, and Italian were sometimes perceived to be word-for-word translations of the German text, which is also reflected in the average scores in \Cref{fig:trans_eval_compact}. 
Other translation errors were cultural; for example, certain Turkish and Polish translations did not sound native to the evaluators due to information about school systems (e.g. referring to a student in ``13\textsuperscript{th} grade'' when, according to the evaluator, Polish school systems do not have a 13\textsuperscript{th} grade) or social services (e.g. ``Betreutes Wohnen,'' or ``assisted living'') which do not necessarily have direct equivalents outside the German context. 

GPT-4.1 was instructed to keep the original number of annotations in the translations without interference. 
After translating each document, we automatically validated the number of annotations in the generated translation compared to the original one in German. 
Invalid translations were regenerated with a maximum number of 10 retries.
MultiGraSCCo retains the original German annotations with an average preservation rate of 99.76\% across all languages, indicating an almost perfect retention.
\Cref{table:automatic_eval} presents the percentages of annotations preserved per language and overall. 


\paragraph{Mono-, Cross- and Multilingual Experiments}

In the monolingual setting with mmBERT, we observe a strong baseline performance across all languages, illustrated in \Cref{table:monolingual_f1_comparison}. PHI detection is consistently high, with PHI micro F1-scores ranging from 0.90 to 0.95 and solid macro F1-scores across most languages. IPI detection, by contrast, remains more challenging. While all IPI micro F1-scores fall within a narrow range, the corresponding macro F1-scores are notably lower. This gap indicates that minority IPI classes are driving much of the overall error. These latter labels typically reflect more subtle, context-dependent cues, unlike PHI entities that often have explicit surface forms (e.g., names or dates). 

Crosslingually, performance declines when the model is trained only on German and evaluated on other languages (\Cref{table:crosslingual_f1}). This drop is most noticeable in IPIs, with IPI macro F1-scores falling as low as 0.42 for Arabic, compared to 0.55 for Italian. These results highlight the difficulty of transferring low-resource, label-specific knowledge across languages without supervision. PHI categories are more stable under this setup, likely due to their more consistent lexical form. However, the variability in macro F1, from 0.56 (Persian) to 0.70 (English), suggests that mmBERT’s ability to generalize across linguistic boundaries remains uneven. This demonstrates the limits of zero-shot transfer and points to the need for more language-specific annotation, particularly for rare entity types.\\

Adding even small amounts of in-language training data results in performance gains in the multilingual setting (Table~\ref{table:multilingual_f1_micro_macro_summary}). At just 25\% of the target language training data, macro F1 surpasses crosslingual scores and in some cases outperforms the monolingual baseline. As expected, model performance improves steadily as more in-language data becomes available, with training on 100\% of the available data having the strongest results overall. The full multilingual setup achieves the highest macro F1-scores across nearly all languages, achieving micro F1s above 0.86 and macro F1s above 0.79 across all languages. This means improved generalization and label stability, especially for IPI entities. Furthermore, these results show that multilingual training offers an effective compromise: it leverages transfer from high-resource languages such as German while benefiting from even limited annotations in the target language. 

\section{Conclusion}
In conclusion, we introduced MultiGraSCCo, a multilingual anonymization benchmark based on the GraSCCo dataset.
To create this benchmark, we leveraged existing annotations of PHIs and added a new annotation layer focusing on indirect personal identifiers. 
Our Inter-Annotator Agreement of 83\% micro averaged pairwise F1-score (macro 73\%) shows a subjective factor when annotating certain indirect personal categories, highlighting the necessity for further exploration.
Using GPT-4.1, we performed an annotation-preserving and culturally adapting translation into nine languages.

Our qualitative evaluation study shows high average scores for translation quality and annotation adaptation in the target languages when assessed by native speakers from the medical domain. 
This shows the ability of language models such as GPT-4.1 not only to translate annotations of personal information into a wide range of languages from different language families, but also to adapt these annotations to various cultural contexts. 

Moreover, our monolingual and crosslingual baseline experiments for automatically detecting personal information using the translations show promising results in all languages. Finally, our multilingual experiments demonstrate that even limited in-language supervision substantially improves model performance, especially for complex IPI labels, highlighting the effectiveness of multilingual training in low-resource settings. This opens new doors for privacy research in under-explored languages by providing high-quality data for improving, building, and testing anonymization systems. 

\section*{Acknowledgements}
We thank the anonymous reviewers for their constructive feedback on this paper. This research was supported by the Federal Ministry of Research, Technology and Space (BMFTR) through the project Medinym (16KISA006 and 16KISA007) and the grant BIFOLD25B.

\section*{Limitations}

The GraSCCo dataset we translate in this work contains several distinctive features which were added by its authors for the purposes of their research. 
In order to challenge de-identification models, some patients and doctors were given first names and surnames based on medical terminology, such as ``Disease'' and ``Alzheimer.'' 
Additionally, some minor errors are present; for example, the patient's and doctor's names are sometimes identical. 
While GPT-4.1 eliminated some of these features in translation, others may remain in MultiGraSCCo.

The original GraSCCo dataset includes a high number of abbreviations as well as typos.
These abbreviations were expanded and typos corrected during preprocessing in order to ensure high-quality translations. As such, this feature of the original GraSCCo data is no longer present in the translations. 
Moreover, we are aware that clinical texts may be written differently in different languages and places, with varying norms, specialized vocabularies, and other distinct linguistic features.
Due to a lack of real, publically available examples which showcase the diverse writing styles and features of clinical documents in each target language, our translated corpus may not follow linguistic or stylistic features unique to the target languages and contexts.
Therefore, the task of replicating more realistic clinical texts in the target languages is left for future work.

In our experiments, we evaluate the performance of the models in detecting personal information on a sample of the translations. 
The performance and usability of these models should still be tested on real patient data in each language.

\section*{Ethics Statement}

No real personal data was processed during the course of this work. 

\section{Bibliographical References}
\label{sec:reference}
\bibliographystyle{lrec2026-natbib}
\bibliography{custom}

\label{lr:ref}

\appendix

\section{IPI Annotation}

\subsection{Annotators}
The annotation team was comprised of two female student assistants with similar cultural backgrounds. 
Both annotators are native German speakers.
One annotator has medical background and is currently studying Business Information Systems, and the other one studies Psychology. 
Both annotators were compensated as part of their regular research assistant roles.

\section{Evaluation of the Automatic Typo Correction and Abbreviation Expansion}
\label{sec:preprocess_eval_sec}

To evaluate this preprocessing step, two annotators inspected 15 documents and counted the number of errors introduced by the model, including missing or inaccurately corrected typos, incorrectly expanded abbreviations, and missing tags. 
For each dimension, we calculate the percentage of errors from the overall number of typos, abbreviations, and tags in the text and report the average percentage of each dimension. 
Our annotators did not detect any typo or tag errors in the generated texts. 
Table~\ref{tab:preproc_eval} summarizes the results of this evaluation.

\begin{table}[H]
\small
\centering
\begin{tabular}{@{}llll@{}}
\toprule
\textbf{} & \textbf{Typos} & \textbf{Abbreviations} & \textbf{Tags} \\ \midrule
Annotator 1        & 0\%            & 4\%                    & 0\%           \\
Annotator 2        & 0\%            & 4\%                    & 0\%           \\ \bottomrule
\end{tabular}
\caption{Average percentages of errors made by the model from the overall number of typos, abbreviations, and tags in each text. 0\% average percentage means that the model processed all of the corresponding data correctly without any errors.}
\label{tab:preproc_eval}
\end{table}

\section{Automatic Translation}
\label{sec:aut_trans}

\begin{figure*}[t]
\centering
\begin{tcolorbox}[width=\textwidth, colback=blue!5!white, colframe=gray!75!black, title=Preprocessing Prompt]
\small

The following text will be translated into different languages. In order to improve the quality of the translations, expand the abbreviations and correct the typos listed below, based on their spans in the following medical text in German. \\
Do not change anything else in the text or rewrite its structure. \\
Only generate the text with the expanded abbreviations and corrected typos. \\
Process only the given lists of typos and abbreviations and do not change anything else in the text. \\
Expand each marked abbreviation, also for units and titles. \\
Note that some words are annotated as both, a typo and abbreviation, make sure to process both in that case. 
Keep the annotation tags that are before and after annotated entities in the generated text, e.g. <NAME\_PATIENT>Baastrup Asger</NAME\_PATIENT>. \\

Abbreviations detected: \{annotated\_abbreviations\}\\

Typos detected: \{annotated\_typos\}\\

Text: \{text\_to\_process\}

\end{tcolorbox}
\caption{The final preprocessing prompt to correct typos and expand abbreviations in the original GraSCCo documents with GPT-4.1.}
\label{fig:preprocess_prompt}
\end{figure*}

\Cref{fig:translation_prompt} presents the final prompt we used for the automatic translation. 
The placeholders \textit{\{source\_language\}}, \textit{\{target\_language\}}, and \textit{\{target\_country\}} were replaced for each language pair accordingly. 
We also specify a target country for each target language to avoid formatting inconsistencies. 
For English we specify USA as the target country, Syria for Arabic, Iran for Persian, Ukraine for Ukrainian, Russia for Russian, France for French, Italy for Italian, Türkiye for Turkish, and finally Poland for Polish. 
\begin{figure*}[t]
\centering
\begin{tcolorbox}[width=\textwidth, colback=blue!5!white, colframe=gray!75!black, title=Translation Prompt]
\small
Translate the following synthetic medical text from \{source\_language\} to \{target\_language\} as spoken in \{target\_country\}. \\
- Keep all <SPLIT> tags exactly as they are. \\
- Translate text inside and outside of any tags such as <NAME\_PATIENT>, <DATE> etc., but do not remove or alter the tags themselves. \\
- Named entities (e.g., patient names, hospitals, cities, locations) must be translated and replaced with natural-sounding random \{target\_language\} equivalents. For example, replace a city mentioned in the original text with a randomly chosen \{target\_language\} city name, and replace a person name mentioned in the original text with a randomly chosen \{target\_language\} name in the translation. Use different random names each time.\\
- Make sure that the adapted entities in the translation such as street names, hospital names, city names etc. exist and are consistent with each other. For example, make sure that the mentioned street or hospital exists in the mentioned city, and that the cities are consistent with the original text, e.g. in terms of distance from each other, in case there are multiple city names mentioned in the original text. \\
- Do not keep the original forms of names or locations (e.g. cities, countries or institutions) and do not use placeholder names (e.g., "Mustermann," "Musterstadt," "John Doe," "Anytown" etc.)\\
- Note that the original texts in \{source\_language\} have names that might not be real names (e.g. Alzheimer), do not reproduce this behavior in the translation, but rather use any real name in \{target\_language\}, even if it does not sound similar to the name in the original text. \\
- Culturally adapt titles (e.g. medical titles such as doctor, professor etc.) in the translation and do not only translate them literally.  \\
- Use the typical format in \{target\_country\} when adapting zip codes and dates in the translation.\\
- Preserve medical terminology with accurate equivalents in \{target\_language\}.\\
- The translation should read fluently and naturally in \{target\_language\}, while respecting the original meaning and structure.\\
- Do not rewrite the structure of the original text.\\
- When translating academic titles, render them idiomatically in \{target\_language\}.\\
- Ensure that sentences are grammatically correct and idiomatic in \{target\_language\}.\\
- Only generate the translation without any further explanations or chitchat.

\end{tcolorbox}
\caption{The final translation prompt and instructions used to translate the clinical documents with GPT-4.1.}
\label{fig:translation_prompt}
\end{figure*}

\section{Inter-Rater Agreement (IRA) for the Translation Evaluation}
\label{sec:ira_translations}
We used Mean Absolute Difference (MAD) to calculate IRA across all rating pairs in all languages and all dimensions.
We calculate the absolute distance between each rating pairs and average it for each dimension across all languages. 
\Cref{table:ira_table} shows detailed analyses about the IRA per dimension. 

\begin{table*}
\small
\centering
\begin{tabular}{lccc}
\toprule
\textbf{Dimension} & \textbf{MAD} & \textbf{Perfect Agreement (\%)} & \textbf{Within One Point (\%)} \\ 
\midrule
General Quality            & 0.667 & 42.22 & 93.33 \\
Grammar                    & 0.711 & 40.00 & 88.89 \\
Tag Structure              & 0.556 & 51.11 & 95.56 \\
Entity Translation Quality & 0.489 & 55.56 & 95.56 \\
\midrule
\textbf{Average}           & \textbf{0.606} & \textbf{47.22} & \textbf{93.34} \\
\bottomrule
\end{tabular}
\caption{Detailed information about Inter-Rater Agreement (IRA) in the translation evaluation study per dimension and on average. The \textbf{Perfect Agreement (\%)} column shows the percentage of rating pairs with the exact same rating. The column \textbf{Within One Point (\%)} shows the percentage of rating pairs which differ by one point at most in the rating scale.}
\label{table:ira_table}
\end{table*}

\section{Additional Training Details}

We fine-tuned Multilingual ModernBERT separately for each language using a single NVIDIA A100 GPU. The data was split at the sentence level including stratified sampling to maintain comparable PHI and IPI label distributions. All hyperparameters were optimized via grid search and early stopping after two epochs' patience. Language-specific hyperparameters are as follows:

\begin{itemize}
    \item Arabic (ar): learning rate 5e-5, batch size 8, epochs 8, weight decay 0.1.
    \item English: learning rate 5e-5, batch size 8, epochs 8, weight decay 0.1.
    \item German: learning rate 5e-5, batch size 8, epochs 8, weight decay 0.0.
    \item French, Persian, Italian, Polish, Russian, Turkish, and Ukrainian: learning rate 3e-5, batch size 8, epochs 6, weight decay 0.0.
\end{itemize}


\begin{table*}
\small
\centering
\begin{tabular}{l|cccccc}
\hline
\textbf{Language} & \textbf{All Micro F1} & \textbf{All Macro F1} & \textbf{PHI Micro F1} & \textbf{PHI Macro F1} & \textbf{IPI Micro F1} & \textbf{IPI Macro F1} \\ \hline
English   & 0.843 \textcolor{gray}{\scriptsize ± 0.009} & 0.744 \textcolor{gray}{\scriptsize ± 0.033} & 0.910 \textcolor{gray}{\scriptsize ± 0.017} & 0.836 \textcolor{gray}{\scriptsize ± 0.041} & 0.770 \textcolor{gray}{\scriptsize ± 0.012} & 0.604 \textcolor{gray}{\scriptsize ± 0.044} \\
French    & 0.833 \textcolor{gray}{\scriptsize ± 0.014} & 0.770 \textcolor{gray}{\scriptsize ± 0.030} & 0.927 \textcolor{gray}{\scriptsize ± 0.009} & 0.891 \textcolor{gray}{\scriptsize ± 0.020} & 0.734 \textcolor{gray}{\scriptsize ± 0.020} & 0.589 \textcolor{gray}{\scriptsize ± 0.065} \\
Arabic    & 0.843 \textcolor{gray}{\scriptsize ± 0.016} & 0.776 \textcolor{gray}{\scriptsize ± 0.018} & 0.936 \textcolor{gray}{\scriptsize ± 0.010} & 0.884 \textcolor{gray}{\scriptsize ± 0.037} & 0.745 \textcolor{gray}{\scriptsize ± 0.032} & 0.614 \textcolor{gray}{\scriptsize ± 0.034} \\
Persian   & 0.827 \textcolor{gray}{\scriptsize ± 0.009} & 0.730 \textcolor{gray}{\scriptsize ± 0.029} & 0.916 \textcolor{gray}{\scriptsize ± 0.009} & 0.829 \textcolor{gray}{\scriptsize ± 0.026} & 0.736 \textcolor{gray}{\scriptsize ± 0.018} & 0.581 \textcolor{gray}{\scriptsize ± 0.039} \\
Italian   & 0.855 \textcolor{gray}{\scriptsize ± 0.014} & 0.793 \textcolor{gray}{\scriptsize ± 0.023} & 0.933 \textcolor{gray}{\scriptsize ± 0.011} & 0.893 \textcolor{gray}{\scriptsize ± 0.023} & 0.767 \textcolor{gray}{\scriptsize ± 0.018} & 0.643 \textcolor{gray}{\scriptsize ± 0.042} \\
Polish    & 0.841 \textcolor{gray}{\scriptsize ± 0.008} & 0.707 \textcolor{gray}{\scriptsize ± 0.054} & 0.906 \textcolor{gray}{\scriptsize ± 0.011} & 0.761 \textcolor{gray}{\scriptsize ± 0.073} & 0.778 \textcolor{gray}{\scriptsize ± 0.022} & 0.628 \textcolor{gray}{\scriptsize ± 0.059} \\
Russian   & 0.839 \textcolor{gray}{\scriptsize ± 0.009} & 0.739 \textcolor{gray}{\scriptsize ± 0.025} & 0.907 \textcolor{gray}{\scriptsize ± 0.011} & 0.819 \textcolor{gray}{\scriptsize ± 0.026} & 0.773 \textcolor{gray}{\scriptsize ± 0.014} & 0.618 \textcolor{gray}{\scriptsize ± 0.039} \\
Ukrainian & 0.841 \textcolor{gray}{\scriptsize ± 0.009} & 0.745 \textcolor{gray}{\scriptsize ± 0.025} & 0.921 \textcolor{gray}{\scriptsize ± 0.014} & 0.852 \textcolor{gray}{\scriptsize ± 0.038} & 0.759 \textcolor{gray}{\scriptsize ± 0.021} & 0.585 \textcolor{gray}{\scriptsize ± 0.012} \\
Turkish   & 0.830 \textcolor{gray}{\scriptsize ± 0.013} & 0.730 \textcolor{gray}{\scriptsize ± 0.032} & 0.899 \textcolor{gray}{\scriptsize ± 0.010} & 0.830 \textcolor{gray}{\scriptsize ± 0.040} & 0.757 \textcolor{gray}{\scriptsize ± 0.020} & 0.579 \textcolor{gray}{\scriptsize ± 0.051} \\ \hline
\end{tabular}
\caption{Multilingual experiment results at 25\% training data per language (\textcolor{gray}{± SD}).}
\label{table:multilingual_f1_025}
\end{table*}

\begin{table*}
\small
\centering
\begin{tabular}{l|cccccc}
\hline
\textbf{Language} & \textbf{All Micro F1} & \textbf{All Macro F1} & \textbf{PHI Micro F1} & \textbf{PHI Macro F1} & \textbf{IPI Micro F1} & \textbf{IPI Macro F1} \\ \hline
English   & 0.851 \textcolor{gray}{\scriptsize ± 0.015} & 0.763 \textcolor{gray}{\scriptsize ± 0.017} & 0.921 \textcolor{gray}{\scriptsize ± 0.013} & 0.858 \textcolor{gray}{\scriptsize ± 0.027} & 0.775 \textcolor{gray}{\scriptsize ± 0.018} & 0.621 \textcolor{gray}{\scriptsize ± 0.029} \\
French    & 0.848 \textcolor{gray}{\scriptsize ± 0.012} & 0.790 \textcolor{gray}{\scriptsize ± 0.032} & 0.932 \textcolor{gray}{\scriptsize ± 0.018} & 0.886 \textcolor{gray}{\scriptsize ± 0.039} & 0.759 \textcolor{gray}{\scriptsize ± 0.031} & 0.646 \textcolor{gray}{\scriptsize ± 0.062} \\
Arabic    & 0.851 \textcolor{gray}{\scriptsize ± 0.008} & 0.781 \textcolor{gray}{\scriptsize ± 0.014} & 0.936 \textcolor{gray}{\scriptsize ± 0.010} & 0.880 \textcolor{gray}{\scriptsize ± 0.015} & 0.759 \textcolor{gray}{\scriptsize ± 0.019} & 0.631 \textcolor{gray}{\scriptsize ± 0.039} \\
Persian   & 0.843 \textcolor{gray}{\scriptsize ± 0.008} & 0.746 \textcolor{gray}{\scriptsize ± 0.017} & 0.928 \textcolor{gray}{\scriptsize ± 0.010} & 0.861 \textcolor{gray}{\scriptsize ± 0.020} & 0.754 \textcolor{gray}{\scriptsize ± 0.018} & 0.574 \textcolor{gray}{\scriptsize ± 0.027} \\
Italian   & 0.862 \textcolor{gray}{\scriptsize ± 0.012} & 0.796 \textcolor{gray}{\scriptsize ± 0.028} & 0.935 \textcolor{gray}{\scriptsize ± 0.009} & 0.889 \textcolor{gray}{\scriptsize ± 0.029} & 0.779 \textcolor{gray}{\scriptsize ± 0.025} & 0.655 \textcolor{gray}{\scriptsize ± 0.031} \\
Polish    & 0.861 \textcolor{gray}{\scriptsize ± 0.013} & 0.720 \textcolor{gray}{\scriptsize ± 0.033} & 0.919 \textcolor{gray}{\scriptsize ± 0.006} & 0.770 \textcolor{gray}{\scriptsize ± 0.069} & 0.805 \textcolor{gray}{\scriptsize ± 0.027} & 0.648 \textcolor{gray}{\scriptsize ± 0.071} \\
Russian   & 0.845 \textcolor{gray}{\scriptsize ± 0.006} & 0.746 \textcolor{gray}{\scriptsize ± 0.023} & 0.920 \textcolor{gray}{\scriptsize ± 0.015} & 0.853 \textcolor{gray}{\scriptsize ± 0.043} & 0.771 \textcolor{gray}{\scriptsize ± 0.011} & 0.587 \textcolor{gray}{\scriptsize ± 0.027} \\
Ukrainian & 0.856 \textcolor{gray}{\scriptsize ± 0.007} & 0.785 \textcolor{gray}{\scriptsize ± 0.015} & 0.931 \textcolor{gray}{\scriptsize ± 0.009} & 0.882 \textcolor{gray}{\scriptsize ± 0.018} & 0.780 \textcolor{gray}{\scriptsize ± 0.017} & 0.639 \textcolor{gray}{\scriptsize ± 0.019} \\
Turkish   & 0.847 \textcolor{gray}{\scriptsize ± 0.014} & 0.776 \textcolor{gray}{\scriptsize ± 0.047} & 0.919 \textcolor{gray}{\scriptsize ± 0.010} & 0.869 \textcolor{gray}{\scriptsize ± 0.045} & 0.772 \textcolor{gray}{\scriptsize ± 0.031} & 0.636 \textcolor{gray}{\scriptsize ± 0.090} \\ \hline
\end{tabular}
\caption{Multilingual experiment results at 50\% training data per language (\textcolor{gray}{± SD}).}
\label{table:multilingual_f1_050}
\end{table*}

\begin{table*}
\small
\centering
\begin{tabular}{l|cccccc}
\hline
\textbf{Language} & \textbf{All Micro F1} & \textbf{All Macro F1} & \textbf{PHI Micro F1} & \textbf{PHI Macro F1} & \textbf{IPI Micro F1} & \textbf{IPI Macro F1} \\ \hline
English   & 0.862 \textcolor{gray}{\scriptsize ± 0.009} & 0.788 \textcolor{gray}{\scriptsize ± 0.033} & 0.931 \textcolor{gray}{\scriptsize ± 0.002} & 0.874 \textcolor{gray}{\scriptsize ± 0.017} & 0.788 \textcolor{gray}{\scriptsize ± 0.015} & 0.659 \textcolor{gray}{\scriptsize ± 0.064} \\
French    & 0.853 \textcolor{gray}{\scriptsize ± 0.013} & 0.800 \textcolor{gray}{\scriptsize ± 0.023} & 0.935 \textcolor{gray}{\scriptsize ± 0.015} & 0.901 \textcolor{gray}{\scriptsize ± 0.025} & 0.766 \textcolor{gray}{\scriptsize ± 0.011} & 0.648 \textcolor{gray}{\scriptsize ± 0.028} \\
Arabic    & 0.857 \textcolor{gray}{\scriptsize ± 0.009} & 0.786 \textcolor{gray}{\scriptsize ± 0.027} & 0.943 \textcolor{gray}{\scriptsize ± 0.009} & 0.887 \textcolor{gray}{\scriptsize ± 0.037} & 0.764 \textcolor{gray}{\scriptsize ± 0.018} & 0.635 \textcolor{gray}{\scriptsize ± 0.063} \\
Persian   & 0.856 \textcolor{gray}{\scriptsize ± 0.015} & 0.766 \textcolor{gray}{\scriptsize ± 0.033} & 0.940 \textcolor{gray}{\scriptsize ± 0.009} & 0.863 \textcolor{gray}{\scriptsize ± 0.035} & 0.769 \textcolor{gray}{\scriptsize ± 0.028} & 0.621 \textcolor{gray}{\scriptsize ± 0.045} \\
Italian   & 0.869 \textcolor{gray}{\scriptsize ± 0.006} & 0.829 \textcolor{gray}{\scriptsize ± 0.025} & 0.940 \textcolor{gray}{\scriptsize ± 0.007} & 0.908 \textcolor{gray}{\scriptsize ± 0.035} & 0.788 \textcolor{gray}{\scriptsize ± 0.014} & 0.710 \textcolor{gray}{\scriptsize ± 0.049} \\
Polish    & 0.865 \textcolor{gray}{\scriptsize ± 0.006} & 0.782 \textcolor{gray}{\scriptsize ± 0.046} & 0.927 \textcolor{gray}{\scriptsize ± 0.012} & 0.855 \textcolor{gray}{\scriptsize ± 0.057} & 0.807 \textcolor{gray}{\scriptsize ± 0.021} & 0.684 \textcolor{gray}{\scriptsize ± 0.046} \\
Russian   & 0.868 \textcolor{gray}{\scriptsize ± 0.010} & 0.803 \textcolor{gray}{\scriptsize ± 0.027} & 0.935 \textcolor{gray}{\scriptsize ± 0.013} & 0.895 \textcolor{gray}{\scriptsize ± 0.031} & 0.802 \textcolor{gray}{\scriptsize ± 0.017} & 0.665 \textcolor{gray}{\scriptsize ± 0.039} \\
Ukrainian & 0.858 \textcolor{gray}{\scriptsize ± 0.014} & 0.786 \textcolor{gray}{\scriptsize ± 0.018} & 0.933 \textcolor{gray}{\scriptsize ± 0.010} & 0.879 \textcolor{gray}{\scriptsize ± 0.028} & 0.782 \textcolor{gray}{\scriptsize ± 0.019} & 0.646 \textcolor{gray}{\scriptsize ± 0.025} \\
Turkish   & 0.856 \textcolor{gray}{\scriptsize ± 0.014} & 0.782 \textcolor{gray}{\scriptsize ± 0.028} & 0.928 \textcolor{gray}{\scriptsize ± 0.011} & 0.876 \textcolor{gray}{\scriptsize ± 0.044} & 0.780 \textcolor{gray}{\scriptsize ± 0.024} & 0.640 \textcolor{gray}{\scriptsize ± 0.066} \\ \hline
\end{tabular}
\caption{Multilingual experiment results at 75\% training data per language (\textcolor{gray}{± SD}).}
\label{table:multilingual_f1_075}
\end{table*}

\begin{table*}
\small
\centering
\begin{tabular}{l|cccccc}
\hline
\textbf{Language} & \textbf{All Micro F1} & \textbf{All Macro F1} & \textbf{PHI Micro F1} & \textbf{PHI Macro F1} & \textbf{IPI Micro F1} & \textbf{IPI Macro F1} \\ \hline
English   & 0.867\textcolor{gray}{\scriptsize ± 0.011} & 0.795\textcolor{gray}{\scriptsize ± 0.017} & 0.936\textcolor{gray}{\scriptsize ± 0.006} & 0.872\textcolor{gray}{\scriptsize ± 0.015} & 0.792\textcolor{gray}{\scriptsize ± 0.020} & 0.679\textcolor{gray}{\scriptsize ± 0.046} \\
French    & 0.863\textcolor{gray}{\scriptsize ± 0.007} & 0.812\textcolor{gray}{\scriptsize ± 0.013} & 0.944\textcolor{gray}{\scriptsize ± 0.010} & 0.919\textcolor{gray}{\scriptsize ± 0.030} & 0.777\textcolor{gray}{\scriptsize ± 0.013} & 0.653\textcolor{gray}{\scriptsize ± 0.050} \\
Arabic    & 0.862\textcolor{gray}{\scriptsize ± 0.019} & 0.763\textcolor{gray}{\scriptsize ± 0.037} & 0.943\textcolor{gray}{\scriptsize ± 0.009} & 0.868\textcolor{gray}{\scriptsize ± 0.024} & 0.774\textcolor{gray}{\scriptsize ± 0.037} & 0.607\textcolor{gray}{\scriptsize ± 0.069} \\
Persian   & 0.865\textcolor{gray}{\scriptsize ± 0.007} & 0.767\textcolor{gray}{\scriptsize ± 0.019} & 0.942\textcolor{gray}{\scriptsize ± 0.008} & 0.867\textcolor{gray}{\scriptsize ± 0.032} & 0.784\textcolor{gray}{\scriptsize ± 0.020} & 0.615\textcolor{gray}{\scriptsize ± 0.022} \\
Italian   & 0.879\textcolor{gray}{\scriptsize ± 0.010} & 0.834\textcolor{gray}{\scriptsize ± 0.012} & 0.950\textcolor{gray}{\scriptsize ± 0.006} & 0.924\textcolor{gray}{\scriptsize ± 0.019} & 0.800\textcolor{gray}{\scriptsize ± 0.017} & 0.699\textcolor{gray}{\scriptsize ± 0.023} \\
Polish    & 0.874\textcolor{gray}{\scriptsize ± 0.010} & 0.783\textcolor{gray}{\scriptsize ± 0.030} & 0.932\textcolor{gray}{\scriptsize ± 0.010} & 0.866\textcolor{gray}{\scriptsize ± 0.038} & 0.818\textcolor{gray}{\scriptsize ± 0.021} & 0.673\textcolor{gray}{\scriptsize ± 0.036} \\
Russian   & 0.867\textcolor{gray}{\scriptsize ± 0.004} & 0.806\textcolor{gray}{\scriptsize ± 0.022} & 0.936\textcolor{gray}{\scriptsize ± 0.012} & 0.899\textcolor{gray}{\scriptsize ± 0.035} & 0.801\textcolor{gray}{\scriptsize ± 0.006} & 0.667\textcolor{gray}{\scriptsize ± 0.025} \\
Ukrainian & 0.867\textcolor{gray}{\scriptsize ± 0.013} & 0.799\textcolor{gray}{\scriptsize ± 0.029} & 0.935\textcolor{gray}{\scriptsize ± 0.008} & 0.885\textcolor{gray}{\scriptsize ± 0.025} & 0.797\textcolor{gray}{\scriptsize ± 0.024} & 0.670\textcolor{gray}{\scriptsize ± 0.037} \\
Turkish   & 0.867\textcolor{gray}{\scriptsize ± 0.012} & 0.817\textcolor{gray}{\scriptsize ± 0.035} & 0.940\textcolor{gray}{\scriptsize ± 0.007} & 0.909\textcolor{gray}{\scriptsize ± 0.026} & 0.790\textcolor{gray}{\scriptsize ± 0.022} & 0.677\textcolor{gray}{\scriptsize ± 0.055} \\
\hline
\end{tabular}
\caption{Multilingual experiment results at 100\% training data per language (\textcolor{gray}{± SD}).}
\label{table:multilingual_f1_100}
\end{table*}

\end{document}